# 2018 Low-Power Image Recognition Challenge

Sergei Alyamkin, Matthew Ardi, Achille Brighton, Alexander C. Berg, Yiran Chen, Hsin-Pai Cheng,
Bo Chen, Zichen Fan, Chen Feng, Bo Fu, Kent Gauen, Jongkook Go, Alexander Goncharenko, Xuyang Guo,
Hong Hanh Nguyen, Andrew Howard, Yuanjun Huang, Donghyun Kang, Jaeyoun Kim, Alexander Kondratyev,
Seungjae Lee, Suwoong Lee, Junhyeok Lee, Zhiyu Liang, Xin Liu, Juzheng Liu, Zichao Li, Yang Lu,
Yung-Hsiang Lu, Deeptanshu Malik, Eunbyung Park, Denis Repin, Tao Sheng, Liang Shen, Fei Sun,
David Svitov, George K. Thiruvathukal, Baiwu Zhang, Jingchi Zhang, Xiaopeng Zhang, Shaojie Zhuo

*Abstract*

The Low-Power Image Recognition Challenge (LPIRC, https://rebootingcomputing.ieee.org/lpirc) is an annual competition started in 2015. The competition identifies the best technologies that can classify and detect objects in images efficiently (short execution time and low energy consumption) and accurately (high precision). Over the four years, the winners' scores have improved more than 24 times. As computer vision is widely used in many battery-powered systems (such as drones and mobile phones), the need for low-power computer vision will become increasingly important. This paper summarizes LPIRC 2018 by describing the three different tracks and the winners' solutions.

## 1. Introduction

Competitions are an effective way of promoting innovation and system integration. "Grand Challenges" can push the boundaries of technologies and open new directions for research and development. The DARPA Grand Challenge in 2004 opened the era of autonomous vehicles. Since 2010, the ImageNet Large Scale Visual Recognition Challenge (ILSVRC) has become a de facto standard benchmark in computer vision. The Low-Power Image Recognition Challenge (LPIRC) started in 2015 as an annual competition identifying the best system-level solution for detecting objects in images while using as little energy as possible (Lu et al. 2015). Although many competitions are held every year, LPIRC is the only one integrating both image recognition and low power. In LPIRC, a contestants' system is connected to the referee system through an intranet (wired or wireless). *There is no restriction on software or hardware.* The contestant's system issues HTTP GET commands to retrieve images from the referee system and issues HTTP POST commands to send the answers. In the past four years, LPIRC has experimented with different tracks with different rules and restrictions. In 2015, offloading of processing was allowed but only one team participated. In 2016, a track displayed images on a computer screen and a constantant's system used a camera to capture the images but only one team participated in this track. These two tracks are no longer offered.

## 2. Tracks in 2018 LPIRC

Vision systems typically consist of three components: model architecture (such as a neural network), inference engine (such as a series of kernels that evaluates neural network operations efficiently on a specific hardware), and the carrying hardware LPIRC tracks prior to 2018 required system-level improvements that did not distinguish between progress in individual components. In 2018, two new tracks were offered that strategically focused on model architecture and the inference engine.

### 2.1 Track 1: TfLite Model on Mobile Phones

This new track, also known as the On-device Visual Intelligence Competition, focused on model architecture: contestants submit their inference neural network model in TfLite format using Tensorflow (https://www.tensorflow.org/mobile/tflite/). The models were benchmarked on a fixed inference engine (TfLite) and hardware model (Pixel 2 XL phone). The task is ImageNet classification. The submissions should strive to classify correctly as many images as possible given a time budget of 30 ms per image. The submissions were evaluated on a single core with a batch-size of one to mimic realistic use cases on mobile devices. Although the scoring metric is based on inference latency rather than energy consumption, the two are usually correlated when the same benchmarks are used on the same hardware.

Track 1's latency-based metric is critical to accelerating the development of mobile model architectures. Prior to this metric, there was no common, relevant and verifiable metric to evaluate the inference speed of mobile model architectures. Numerous papers characterize inference efficiency using unspecified benchmarking environments or latency on desktop as a proxy. Even the most commonly used metric, MACs (multiply-add count) do not correlate well with inference latency in the real-time (under 30 ms) range (Chen and Gilbert 2018).

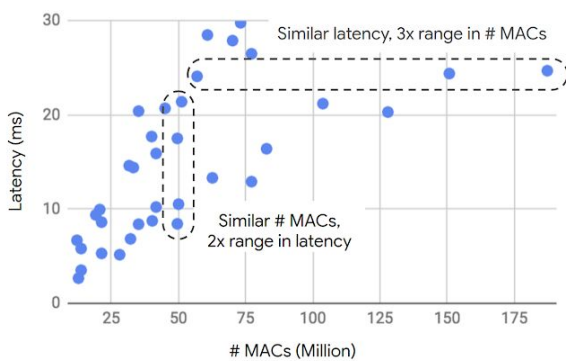

Track 1's metric is the first to measures on-device latency in a realistic use case in the real-time range. In addition, this track provides a benchmarking platform (see Section 5.1) that allows for repeatable measurements and fair comparison of mobile model architectures. The flexibility of submitting just the model and not the full system allows this track to be an online competition. This convenience helps to boost the submission count to 130 models within just 2 weeks.

The submissions also witness excellent quality, establishing the new state-of-the-art in mobile model architecture design. Within the 30 ms latency range, the best of track 1 submissions outperformed the previous state-of-the-art based on quantized MobileNet V1, by 3.1% on the hold-out test set.

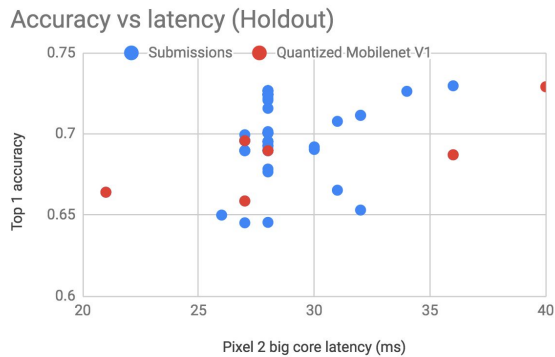

The submissions have demonstrated a considerable amount of diversity. Slightly over half (51.7%) of the solutions are quantized. Most of the architectures (74.1%) are variations of the existing Mobilenet model family, namely quantized V2 (22.4%), quantized V1 (24.1%) and float V2 (27.6%). The predominant dependence on Mobilenets is not surprising, considering their exceptional performance on-device and the convenient support by TfLite. On the other hand, to sustain long-term mobile vision research, track 1 should also serve to reward previously under- or un-explored model architectures. In future installments of LPIRC, track 1 is looking to implement mechanisms for facilitating the discovery of novel architectures and rendering the platform more inclusive to general vision researchers.

### 2.2 Track 2: Caffe 2 and TX 2

LPIRC aims to discover the most energy-efficient solutions for detecting objects in images. For this reason, LPIRC is a system-level competition and contestants must port their solutions to functional systems. To reduce the barrier to entry, Track 2 uses pre-selected software (Caffe 2) and hardware (Nvidia TX2). A software development kit (SDK) is provided for installing the necessary packages. There are two main requirements for Track 2:

1. Participants use Caffe2 (http://caffe2.ai/) to build their image detection systems. For example, the official code of Faster R-CNN is written by Caffe, participants need to convert it to Caffe2.

2. Participants use Nvidia TX2 (http://developer.nvidia.com/embedded/buy/jetson-tx2) as the hardware platforms.

With these conditions, the solutions for Track 2 can be submitted online and evaluated in the organizers' laboratory. The score of track 2 is the ratio between recognition accuracy (mean average precision, mAP) and the total amount of energy consumption (Watt-hour). The energy consumption is measured using a power meter of the power supply to TX2.

Each team has 10 minutes to process all the images. There are total 200 classes of objects in the competition. These classes are the same as ImageNet's. Contestant's training set is not restricted to ImageNet. They can use any datasets to train the model.

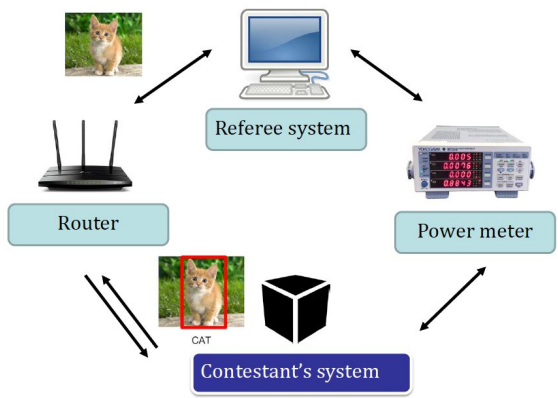

Test images are stored in the referee system and retrieved by TX2 through the router. The results are uploaded to the referee system through the network. After the contestant's system logs in, the power meter starts to measuring the power consumption. The power meter stops after 10 minutes or a contestant's system logs out. This allows a team that finishes processing all images within 10 minutes to reduce their overall energy consumption.

**2.3 Track 3: Onsite, No Restriction**

The original track, always offered since 2015, was also available in 2018. This track has no restriction in hardware or software and gives contestants the most flexibility. Additional information about Track 3 can be obtained from prior publications (Gauen et al. 2017), (Gauen et al. 2018), (Lu, Berg, and Chen 2018).

**2.4 Training Data**

LPIRC uses the same training images as ImageNet Large Scale Visual Recognition Challenge (ILSVRC). For Track 1, the data from localization and classification task (LOC-CLS) is used. It consists of around 1.2 million photographs, collected from flickr and other search engines, hand-labeled with the presence or absence of 1000 object categories. Bounding boxes for the objects are available, although Track 1 only considers classification accuracy as a performance metric. For track 2 and 3, the data from object detection is used. It has around 550,000 images and bounding boxes of labels for all 200 categories.

**2.5 Testing Data**

For Track 1, the test data is newly created by using the ILSVRC image crawling tool to collect 100 images for each category from Flickr. When crawling the images, synonyms (e.g. "house finch, linnet, Carpodacus mexicanus") are used in the searching to ensure that the images are relevant to the corresponding categories. The competition uses only the images uploaded after June 2017 to avoid duplications with the test images used in the previous years' LPIRC. Thumbnail images are used to remove duplicates by resizing the images to 30 x 30, and calculating the L2-norm of differences with images in the previous dataset. Ten representative images are manually chosen from each category. Representative images refer to the images that can be identified as one of the 1000 ImageNet classification categories without ambiguity. The following photos show examples.

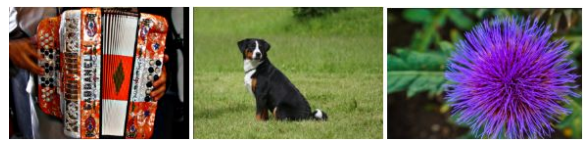

Accordion     Appenzeller     Artichoke

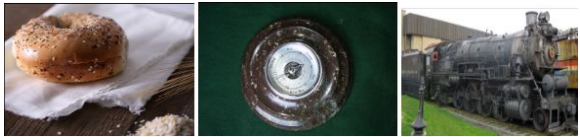

| Bagel | Barometer | Steam Locomotive |

Tracks 2 and 3 also use the ILSVRC tool to crawl images from Flickr with the context query phrases (e.g., cat in the kitchen) instead of the label names (e.g., cat) as the keywords. As a result, the tool can obtain images with various types of objects. The annotations (bounding boxes) are created by using Amazon Mechanical Turk (MTurk) jobs. The tracks use 20,000 test images.

### 2.6 Other Datasets for Image Recognition

ImageNet (provided by ILSVRC) is one of many popular datasets for evaluating computer vision. Other commonly used datasets include PASCAL VOC, COCO (Common Objects in COntext), SUN (Scene UNderstanding), INRIA Pedestrian Dataset, KITTI Vision Benchmark Suite, and Caltech Pedestrian Dataset. Over time the image datasets have improved in two primary ways. First, the quality of the image annotations has significantly improved due to more sophisticated methods for crowdsourcing. Second, the variety of the dataset has increased, in both content and annotations, by increasing the number of classes. The figure below shows PASCAL VOC (left) and ImageNet (right). Datasets and their corresponding competitions have yielded great improvements in computer vision technology.

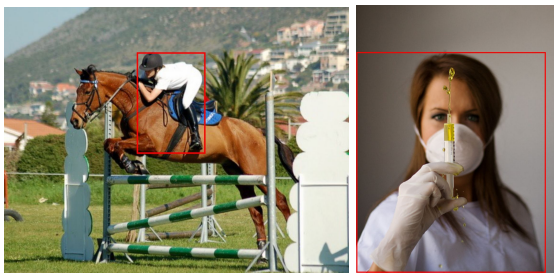

PASCAL VOC started its first challenge in 2005 for object detection and classification of four classes. In 2012, the final year of the competition, the PASCAL VOC training and validation datasets consisted of 27,450 detection objects in 11,530 images with 20 classes. From 2009 - 2005 the overall classification AP improved from 0.65 - 0.82 and detection AP improved from 0.28 - 0.41 (Everingham et al. 2015).

The COCO competition continues to be held annually with a primary focus of correctness of solutions. COCO contains over 2.5 million labeled instances in over 382,000 images with 80 common objects for instance segmentation. COCO also contains other types of annotations including Keypoints, Panoptic, and Stuff (Lin et al. 2014) (Caesar et al. 2018). The performance of a model on COCO has improved for bounding-box object detection has improved from 0.373 to 0.525 from 2015 - 2017.

The significant improvements of model accuracy demonstrated by just two of these image processing competitions is a similar goal of LPIRC. In addition to accuracy, LPIRC also considers energy consumption and execution time.

### 3. LPIRC Scores

### 3.1 Object Detection and Mean Average Precision

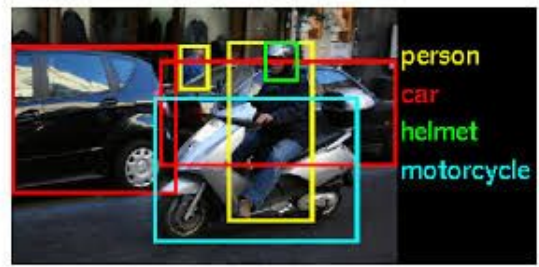

To win Tracks 2 and 3 of LPIRC, a contestant's solution must be able to detect objects in images and mark the objects by bounding boxes. The objects belong to 200 pre-defined categories, such as person, car, airplanes, etc. The image shows an example containing two people, two cars, one motorcycle, and one helmet. The accuracy is measured by Mean Average Precision (mAP). A successful object detection must identify the category correctly and the bounding box must overlap with the correct solution (also called "ground truth") by at least 50%. This is the same definition as ILSVRC. Unlike ILSVRC, however, LPIRC has time limitation: each solution has only 10 minutes to process all images (5,000 in year 2015 and 20,000 since 2016). A superior solution must be able to detect objects and their locations in the images quickly, while consuming as little energy as possible.

### 3.2 Energy Consumption

A system's energy consumption is measured using a high-speed power meter, Yokogawa WT310 Digital Power Meter. It can measure AC or DC and can synchronize with the referee system for the 10-minute time limit.

### 3.3 Score = mAP / Energy

| Year | Accuracy | Energy | Score | Ratio |
|---|---|---|---|---|
| 2015 | 0.02971 | 1.634 | 0.0182 | 1.0 |
| 2016 | 0.03469 | 0.789 | 0.0440 | 2.4 |
| 2017 | 0.24838 | 2.082 | 0.1193 | 6.6 |
| 2018-2 | 0.38981 | 1.540 | 0.2531 | 13.9 |
| 2018-3 | 0.18318 | 0.412 | 0.4446 | 24.4 |

The score of each solution is the ratio of mAP and the energy consumption. The accuracy is measured by the mean average precision; the energy is measured by Watt-hour. The last column shows the progress of the winners' solutions. From 2015 to 2018, the winners' solutions improved by a factor of 24. The row of 2018-2 is the track using Nvidia TX2 and Caffe2. 2018-3 has no restriction of hardware or software.

### 3.4 Track 1 Scores

Track 1 is held for the first time in 2018. The table shows the score of the winner of Track 1. The holdout set is freshly collected for the purpose of the competition.

|  | ImageNet Validation Set | Holdout Set |
|---|---|---|
| Latency | 28.0 | 27.0 |
| Test Metric | 0.64705 | 0.72673 |
| Accuracy on Classified | 0.64705 | 0.72673 |
| Accuracy / Time | 1.08 E-06 | 2.22 E-06 |
| # Classified | 20000 | 10927 |

**Latency:** Latency (ms) is single-threaded, non-batched runtime measured on a single Pixel 2 big core of classifying one image.
**Test metric (main metric):** is the total number of images corrected in a wall time of 30 ms ✕ N divided by N, where N is the total number of test images.
**Accuracy on Classified:** is the accuracy in [0, 1] computed based only on the images classified within the wall-time.
**# Classified:** is the number of images classified within the wall-time.
**Accuracy/Time:** is the ratio of the accuracy and either the total inference time or the wall-time, whichever is longer.

Track 1 received 128 submissions. The figure below represents a total of 121 valid submissions (submissions that passed the bazel test and successfully evaluated). 56 submissions received test metric scores between 0.59 and 0.65.

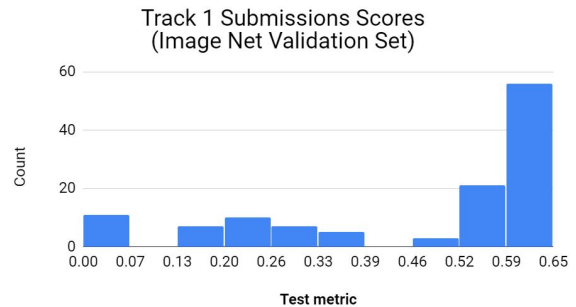

**Mean:** 0.4610    **Median:** 0.5654
**Mode:** 0.6347    **Standard Deviation:** 0.2119

Some duplicate submissions that are successfully evaluated have differences with the "Test Metrics" score due to changes in the evaluation server. These changes are made throughout the progress of the competition. Files that have the same md5 hash information are considered identical/duplicates. Duplicates may be caused by repeating submission done by the participants or re-submission by the same team using different accounts. Files with identical md5 hash information are grouped and only one file with the best "test metric" score is considered to be the unique file among the identical files. If the "test metric" values of all duplicates are equal, only one of the submissions is considered to be unique. After eliminating duplicates, there are 97 unique submissions.

**3.5 Tracks 2 and 3 Scores**

The table shows the top teams in track 2. The winner achieved very high accuracy and finished recognizing all 20,000 images in 5 min. The 2018 winner's final score is twice as high as the 2017 best score.

| Team | mAP | Energy | Score |
|---|---|---|---|
| winner | 0.3898 | 1.3667 | 0.2852 |
| second | 0.1646 | 2.6697 | 0.0616 |
| third | 0.0315 | 1.2348 | 0.0255 |

The table shows the scores for the winners. Two teams' scores are too very close and both teams win the second prize. Also, the top three scores outperform the 2017 winner. Two teams' energy consumption is much lower than previous years' winners'.

| Team | mAP | Energy | Score |
|---|---|---|---|
| winner | 0.1832 | 0.4120 | 0.44462 |
| second | 0.2119 | 0.5338 | 0.39701 |
| second | 0.3753 | 0.9463 | 0.39664 |
| third | 0.2235 | 1.5355 | 0.14556 |

**4. Winners' Solutions**

**4.1 First Prize of Track 1**

The Qualcomm team wins the first prize of Track 1. The competition is driven by the real-world need for accurate real-time image classification using neural network models running on edge devices. Qualcomm provides edge AI platforms with Snapdragon (including the Google Pixel 2 phone used in this competition) so that Smartphone could become real intelligent devices in the near future.

Accurate and fast image recognition on the edge device requires several steps. First, a neural network model needs to be built and trained to identify and classify images (recognizing when a photo is of a dog vs. a cat for example). Then, the model should run as accurate as possible on the actual hardware without latency issues. As most neural networks are trained on a floating-point model, they usually need to be converted to fixed-point in order to efficiently run on edge devices.

For this competition, the model is based on MobileNet V2, but is modified to be quantization-friendly. Although Google's MobileNet models successfully reduce parameter size and computation latency due to the use of separable convolution, directly quantizing a pre-trained MobileNet v2 model can cause huge precision loss. The team analyzes and identifies the root cause of accuracy loss due to quantization in such separable convolution networks, and solved it properly without utilizing quantization-aware re-training.

In separable convolutions, depthwise convolution is applied on each channel independently. However, the min and max values used for weights quantization are taken collectively from all channels. An outlier in one channel may cause a huge quantization loss for the whole model due to an enlarged data range. Without correlation crossing channels, depthwise convolution may be prone to produce all-zero values of weights in one channel. This is commonly observed in both MobileNet v1 and v2 models. All-zero values in one channel means small variance. A large "scale" value for that specific channel would be expected while applying batch normalization transform directly after depthwise convolution. This hurts the representation power of the whole model.

As a solution, the team proposes an effective quantization-friendly separable convolution architecture, where the nonlinear operations (both batch normalization and ReLU6) between depthwise and pointwise convolution layers are all removed, letting the network learn proper weights to handle the batch normalization transform directly. In addition, ReLU6 is replaced with ReLU in all pointwise convolution layers. From various experiments in MobileNet v1 and v2 models, this architecture shows a significant accuracy boost in the 8-bit quantized pipeline.

Fixed-point inferencing while preserving a high level of accuracy is the key to enable deep learning use cases on low power edge devices. The team identifies the industry-wide quantization issue, analyzes the root cause, and solves it on MobielNets efficiently. The quantized modified MobileNet_v2_1.0_128 model can achieve 28 milliseconds per inference with high accuracy (64.7% on ImageNet validation dataset) on a single ARM CPU of Pixel 2. More details are described in the paper "Quantization-friendly separable convolution architecture for MobileNets" (https://arxiv.org/abs/1803.08607).

**4.2 Third Prize of Track 1**

The Expasoft team wins the third prize of Track 1. The team builds a neural network architecture that gives the high accuracy and inference time equal to 30 ms on Google Pixel 2. The team chooses two most promising architectures MobileNet and MobileNet-v2.

Running MobileNet-v1_224 with float-32 on Pixel-2 phone gives 70% accuracy and inference time of 81.5 ms. The team chooses two main directions to accelerate neural network on device: neural network quantization and reducing input image resolution. Both methods lead to accuracy reduction and the team finds trade-off for accuracy – speed relation. Tests of MobileNet and MobileNet-v2 architectures suggest quantizing the neural networks into uint8 data format. The team's evaluation shows final score equal to 68.8% accuracy and 29ms inference time.

| Neural network architecture | Input image resolution | Data type | Accuracy (%) | Google Pixel-2 inference time (ms) | Accuracy/inference time (%/ms) |
|---|---|---|---|---|---|
| mobilenet v1 | 224x224 | float32 | 70.2 | 81.5 | 0.86 |
| mobilenet v1 | 224x224 | uint8 | 65.5 | 68.0 | 0.96 |
| **mobilenet v1** | **128x128** | **uint8** | **64.1** | **28.0** | **2.28** |
| mobilenet v2 | 150x150 | uint8 | 64.4 | 36.6 | 1.75 |
| mobilenet v2 | 132x132 | uint8 | 62.7 | 31.8 | 1.97 |
| mobilenet v2 | 130x130 | uint8 | 59.9 | 31.2 | 1.91 |

Quantization to uint8 allows to reduce inference size from 81ms to 68ms but leads to significant accuracy drops. During standard quantization process in Tensorflow it is required to start from full-precision trained model and learn quantization parameters (min and max values). Instead of joint training of neural network and tuning quantization parameters, the Expasoft team proposes an another approach: tuning quantization parameters using Stochastic Gradient Descent approach with State Through Estimator (Bengio et al. 2013) of gradients of discrete functions (round, clip) without updating weights of the neural networks. Loss function for this process is L2 for embedding layers of full-precision and quantized networks. Mobilenet-v1 224x224 quantized such way shows 65.3% accuracy. Proposed method requires more detailed research but have two significant advantages:
1. method doesn't require labeled data for tuning quantization parameters
2. training goes faster, because no need to train neural network, tune only quantization parameters

**4.3 First Prize of Track 2**

The Seoul National University team wins the first prize in Track 2. The hardware device is Jetson TX2. Optimization is needed to balance speed and accuracy for existing deep learning algorithms originally designed for fast servers. The team discovers that optimization efficiency differs from network to network. For this reason, the team compares various object detection networks on the Jetson TX2 board to find the best solution.

The team compares five modern one-stage object detectors: YOLOv2, YOLOv3, TinyYOLO, SSD, and RetinaNet. Two-stage detectors such as Faster R-CNN are excluded because they are slightly better in accuracy than the one-stage detectors but much slower in speed. The baseline for each network is selected as the authors' published form. For the comparison, the networks are improved with several software techniques briefly introduced as follows.
(1) *Pipelining:* An object detection network can be split into three stages: input image pre-processing, a convolutional network body, and post-processing with output results of the body. Since the network body is typically run on the GPU and the others are on the CPU, they can be executed concurrently by the well-known pipelining technique. (2) *Tucker decomposition:* It is one of the *low-rank approximation* techniques. As described in (Tucker 1966), 2-D Tucker decomposition is used in the network comparison. (3) *Quantization:* For the Jetson TX2 device, only 16-bit quantization is allowed and it is applied to the networks. (4) *Merging batch normalization layer into weights:* Since the batch normalization layer is a linear transformation, it can be integrated into the previous convolutional layer by adjusting the filter weights before running the inference. (5) *Input image size reduction:* Reducing input image size is an easy solution to enhance the network throughput. It was also observed in experiments that the effect of this technique depends on the networks.

By comparing the networks, the team finds that *YOLOv2* outperforms other networks for the on-device object detection with the Jetson TX2. The table shows how much the YOLOv2 network is enhanced by the series of the improvements. Since the total energy consumption is inversely proportional to the network speed, the score (mAP/Wh) can be estimated as mAP x speed. This optimized *YOLOv2* network is selected for the LPIRC Track 2 application.

| Description | mAP(A) | FPS(B) | Score (A x B) | Normalized score |
|---|---|---|---|---|
| Baseline (416x416) | 51.1 | 7.97 | 407 | 1.0 |
| Pipelining | 51.1 | 8.85 | 452 | 1.11 |
| Tucker | 50.2 | 15.1 | 758 | 1.86 |
| Quantization | 50.2 | 19.9 | 999 | 2.45 |
| 256 x 256 | 43.0 | 32.5 | 1640 | 4.03 |
| Batch = 16 | 43.0 | 90.3 | 3880 | 9.54 |

YOLOv2 is tested on the Darknet framework in the experiments and it needs to be translated to Caffe2 framework for Track 2. The team implemente custom Caffe2 operators to support Darknet-specific operators and optimization techniques such as pipelining and 16-bit quantization. Additionally, various batch sizes for the network are tested to find the best batch size for the submission.

Through the steps illustrated above, the estimated score for the YOLOv2 has increased about 9.54 times compared to the baseline and this result surpassed the other object detection networks on the Jetson TX2.

**4.4 Third Prize of Track 2**

The team's members are from Tsinghua University, University of Science and Technology of China, and Nanjing University. The team evaluates several mainstream object detection neural models and picks the most efficient one. The selected model is then fine-tuned with sufficient dataset before being quantized into 16-bit float datatype in order to achieve better power-efficiency and time-efficiency in the exchange of minimal accuracy loss.

The team explores popular object detection neural architectures such as YOLO, RCNN and their variants. Among these architectures, YOLO V3 achieves the best balance between accuracy and computational cost. However, considering the difficulty of engineering implementation in a short amount of time, the team chooses faster RCNN as the base structure and then quantizes the parameters in order to shorten the inference time and reduce power consumption.

Faster RCNN consists of 2 sub-networks--the feature extraction network and the detection network. While the latter doesn't seem to have much space for altering, there're many options for the former, such as VGG and MobileNet. The MobileNet family is known for their much lower computational cost for achieving equivalent accuracy compared with traditional feature extraction architectures. Although MobileNets are reported with good classification results, the mAP for object detection seems low. The team decides to choose VGG-16.

The overview of the software optimization methods can be seen in the figure below. The team reshapes the images on the CPU in order to adjust the image shape to the input shape of the module and then conducts the inference on the GPU. After the inference, the images are reshaped again to obtain the coordinates of the bounding boxes. The team applies three different techniques to accelerate the image recognition process: TensorRT-based inference, 16-bit Quantization, and CPU multithreading.

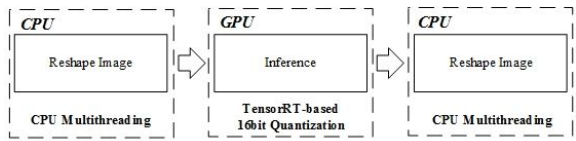

The implements the well-trained Fast-RCNN model used for image recognition on TensorRT. NVIDIA TensorRT (https://developer.nvidia.com/tensorrt) is a high-performance deep learning inference optimizer and runtime that delivers low latency and high-throughput for deep learning inference applications. The inference time becomes much shorter by using TensorRT and therefore saves energy. TensorRT provides quantization operation for the GPU inference engine. The computation latency is shortened because of less floating point arithmetic. In order not to reduce the mAP of the model, the team quantizes the weight to 16-bit while doing inference.

The method reshapes the images before and after the inference process. Since reshaping on the CPU is costly, the method uses CPU multithreading to accelerate the process. The method generates two threads on each CPU core and each thread process one batch of data. TX 2 has 6 CPU cores and the method creates 12 threads. The inference on the GPU can only work in a single thread; thus the method takes the inference as a mutual process and different threads need to compete for the GPU.

Applying multithreading method results in a 1.4x speeding up (the fps without multithreading is 3.17 while the fps with multithreading is 4.41), not as high as expected. The reason is competition for the mutual recourse among the threads and constrains acceleration. Different threads need to wait for the inference process when it's occupied and need to use mutual lock to avoid the hazard.

### 4.5 First and Second Prizes of Track 3

The ETRI and KPST team wins both the first and the second prizes of Track 3. The performance of object detection can be evaluated using accuracy, speed and memory. The accuracy is a common measurement and it has been widely used in comparing different object detection architectures. However, its performance is dependent upon speed and memory as well as accuracy for the given environments and applications. In this competition, the performance is measured using accuracy (mAP) and energy consumption (WH). The accuracy is influenced by the detection architectures, an input resolution and a base network for feature extraction. The more complex structures with a high-resolution image may achieve higher accuracy at higher energy consumption. To obtain a high score, it is necessary to balance accuracy and energy consumption.

The team examines the score function and finds the most important factor is energy consumption. At the same accuracy, the score is higher at lower energy consumption. Moreover, the accuracy is weighted by the processed images ratio within 10 minutes. This means that the detection architecture should be fast and light and the trade-off between accuracy and energy consumption. Accordingly, single stage detectors such as SSD (Liu et al. 2016) and YOLO (Redmon and Farhadi 2017) are considered in the pipeline. The team selects SSD as a detection framework due to its simplicity and stable accuracy among feature extractors (Huang et al. 2017). To obtain the best score, the team performs three optimization steps: 1) Detection structure optimization, 2) Feature extraction optimization, and 3) System and parameters optimization.

For detection structure optimization, in the detection structure, the original SSD is based on VGG and its performance is well balanced in accuracy and speed. Its speed and memory can be improved for low-power and real-time environments. To speed up, the team proposes efficient SSD (eSSD) by adding additional feature extraction layers and prediction layers in SSD. The following table shows the comparison of SSD (Liu et al. 2016), SSDLite (Sandler et al. 2018) and eSSD. In SSD, additional feature extraction is computed by 1x1 conv and 3x3 conv with stride 2 and prediction uses 3x3 conv. The SSDLite replaces all 3x3 conv with depthwise conv and 1x1 conv. The eSSD extracts additional features with depthwise conv and 1x1 conv and predicts classification and bounding box of an object with 1x1 conv. This reduces memory and computational resources.

| Type | Additional feature extraction | Prediction |
|---|---|---|
| SSD | 1x1 conv / 3x3 conv-s2 | 3x3 conv |
| SSDLite | 1x1 conv / 3x3 conv(dw)-s2 / 1x1 conv | 3x3 conv(dw) / 1x1 conv |
| eSSD | 3x3 conv(dw)-s2 / 1x1 conv | 1x1 conv |

The comparison of feature extraction and prediction in SSD, SSDLite and eSSD.

The next table shows accuracy, speed and memory comparison of SSD variants in VOC0712 database. In this experiment, a simple prediction layer such as 3 by 3 or 1 by 1 is applied and an inference time (forward time) is measured in a single Titan XP (Pascal) GPU. SSDLite is more efficient in memory usage than eSSD, but eSSD shows better performance in speed than SSDLite.

| Base Network (300x300) | Feature extraction / Prediction | mAP | Speed (ms) | Model (MB) |
|---|---|---|---|---|
| MobileNetV1 | SSD / 3x3 | 68.6 | 8.05 | 34.8 |
| MobileNetV1 | SSD / 1x1 | 67.8 | 6.19 | 23.1 |
| MobileNetV1 | SSDLite /1x1 | 67.8 | 5.91 | 16.8 |
| MobileNetV1 | eSSD / 1x1 | 67.9 | 5.61 | 17.4 |
| MobileNetV1 ($C$=0.75) | eSSD / 1x1 | 65.8 | 5.20 | 11.1 |
| MobileNetV1 ($C$=0.75) | eSSD(L=5) /1x1 | 65.8 | 4.62 | 10.9 |
| VGG | SSD / 3x3 | 77.7 | 12.43 | 105.2 |

The accuracy (mAP), speed (ms) and memory (MB) for different feature extraction and prediction architectures in VOC 0712 train and VOC 2007 test dataset.

For feature extraction optimization, the base model of feature extractor, MobilNetV1 (Howard et al. 2017) is used and feature extraction layers of MobileNetV1 is optimized in eSSD. To improve memory usage and computational complexity, the team uses 75% weight filter channels ($C$=0.75). Although this drops accuracy, energy consumption is greatly reduced. The team uses five additional layers and modified anchors for a low resolution image. It generates a small number of candidate boxes and improves detection speed. After detection structures are optimized, the team also modifies MobileNetV1 by applying early down-sampling and weight filter channel reduction in earlier layers and trained the base model (MobileNetV1+) from the scratch. All models are trained with ImageNet database and Caffe framework. In training, batch normalization is used and trained weights are merged into final model as introduced in (Fu et al. 2017).

For system and parameters optimization, after training the models, the system is set up and the trained models are ported into NVIDIA TX2. In object detection, multiple duplicate results are obtained and Non-Maximal Suppression (NMS) with thresholding is important. The team tunes NMS process between CPU and GPU to reduce computational complexity and adjust the NMS threshold to decrease result file size for the network bandwidth. Then batch size modification and queuing are applied to maximize speed in detection and to increase the efficiency of network bandwidth. After tuning the speed, to minimize energy consumption, the team uses the low power mode (max-Q mode) in NVIDIA Jetson TX2. The table shows final model specifications in LPIRC 2018 Track 3.

| Models | Accuracy (mAP) | Time (sec) | Energy (WH) | Score |
|---|---|---|---|---|
| eSSD-MobileNetV1+ ($C$=0.75, $I$=160, $th$=0.05, $batch$= 96) | 18.318 | 300 | 0.4119 | 0.4446 (1st) |
| eSSD-MobileNetV1+ ($C$=0.75, 192, $th$=0.05, $batch$=64) | 21.192 | 350 | 0.5338 | 0.3970 (2nd) |

Proposed (eSSD-MobileNetV1+) Model specifications for LPIRC 2018 Track 3. $C$, $I$, $th$ and $batch$ represent channel reduction, input resolution, NMS threshold and batch size, respectively.

### 4.6 Second Prize of Track 3

The Seoul National University team shares the second prize of Track 3 because the scores are very close. The team chooses TX2 board as the hardware platform because of the GPU-based deep learning application. The object detection network is an improve *YOLOv2* network. *The C-GOOD framework (Kang et al. 2018)* is used. Derived from Darknet, this framework helps explore a design space of deep learning algorithms and software optimization options. The team reduces the energy consumption by managing the operating frequencies of the processing elements. The team discovers the optimal combination of CPU and GPU frequencies.

**5. Industry Support for Low-Power Computer Vision**

**5.1 Google Tensorflow Models**

Google provided technical support to Track 1 in two aspects. First, Google open-sources TfLite, an mobile-centric inference engine for Tensorflow models. TfLite encapsulates optimal implementations of standard operations such as convolutions and depthwise-convolutions that shields the contestants from such concerns. This effectively allows people with no background in mobile optimization to participate in the competition. Second, Google provides the mobile benchmarking system and it allows repeatable measurement of the performance metric. The system comes in two flavors: a stand-alone App that the contestants can run on their local phones, and a cloud-based service where the submissions will be automatically dispatched to a Pixel 2 XL phone and benchmarked using the standardized environment. The App (opensourced at https://github.com/tensorflow/tensorflow/tree/master/tensorflow/contrib/lite/java/ovic#measure-on-device-latency) and a validation tool (https://github.com/tensorflow/tensorflow/tree/master/tensorflow/contrib/lite/java/ovic#run-tests) are provided to facilitate rapid development, exploring novel models and catching runtime bugs.
.
Once the model is sufficiently polished, it is submitted to the cloud service, which as a daily submission cap, for refinements and verification.

**5.2 Facebook AI Performance Evaluation Platform**

Machine learning is a rapidly evolving area with many moving parts: new and existing framework enhancements, new hardware solutions, new software backends, and new models. With so many moving parts, it is very difficult to quickly evaluate the performance of a machine learning model. However, such evaluation is vastly important in guiding resource allocation in:
- The development of the frameworks
- The optimization of the software backends
- The selection of the hardware solutions
- The iteration of the machine learning models

Because of this need, Facebook has developed an AI performance evaluation platform (FAI-PEP, open sourced at https://github.com/facebook/FAI-PEP) to provide a unified and standardized AI benchmarking methodology.

FAI-PEP supports Caffe2 and TFLite frameworks, the iOS, Android, linux, and windows operating systems. The FAI-PEP is modularly composed that new frameworks and backends can be added easily. The built-in metrics collected by FAI-PEP are: latency, accuracy, power, and energy. It also supports reporting arbitrary metrics that the user desires to collect. With FAI-PEP, the benchmark runtime condition can be specified precisely, and the ML workload can be benchmarked repeatedly with low variance.

**6. Future Low-Power Computer Vision**

In 2018 CVPR, LPIRC invites three speakers from Google and Facebook sharing their experience building energy-efficient computer vision. More than 100 people attend the workshop. The panel after the speeches answers many attendees' questions. The high participation suggests that there is strong interest, in both academia and industry, to create datasets and common platforms (both hardware and software) for benchmarking different solutions. Readers interested in contributing to future low-power computer vision are encouraged to contact the LPIRC organizers for further discussion.

**7. Conclusion**

This paper explains the three tracks of the 2018 Low-Power Image Recognition Challenge. The winners describe the key improvements in their solutions. As computer vision is widely used in many battery-powered systems (such as drones and mobile phones), the need for low-power computer vision will become increasingly important. The initial success of the novel tracks in 2018 also showcases the advantages of making focused advances on specific components of the vision system, as well as of lowering the bar-of-entry to be inclusive of the general vision and machine learning communities.

**Acknowledgments**

**Authors' Affiliations**

The authors are ordered alphabetically.

Sergei Alyamkin (Expasoft)
Matthew Ardi (Purdue)
Achille Brighton (Google)
Alexander C. Berg (University of North Carolina at Chapel Hill)
Yiran Chen (Duke)
Hsin-Pai Cheng (Duke)
Bo Chen (Google)
Zichen Fan (Tsinghua University)
Chen Feng (Qualcomm)
Bo Fu (Purdue, Google)
Kent Gauen (Purdue)
Jongkook Go (ETRI)
Alexander Goncharenko (Expasoft)
Xuyang Guo (Tsinghua University)
Hong Hanh Nguyen (KPST)
Andrew Howard (Google)
Yuanjun Huang (University of Science and Technology of China)
Donghyun Kang (Seoul National University)
Jaeyoun Kim (Google)
Alexander Kondratyev (Expasoft)
Seungjae Lee (ETRI)
Suwoong Lee (ETRI)
Junhyeok Lee (KPST)
Zhiyu Liang (Qualcomm)
Xin Liu (Duke)
Juzheng Liu (Tsinghua University)
Zichao Li (Nanjing University)
Yang Lu (Facebook)
Yung-Hsiang Lu (Purdue)
Deeptanshu Malik (Purdue)
Eunbyung Park (University of North Carolina at Chapel Hill)
Denis Repin (Expasoft)
Tao Sheng (Qualcomm)
Liang Shen (Qualcomm)
Fei Sun (Facebook)
David Svitov (Expasoft)
George K Thiruvathukal (Loyola University Chicago and Argonne National Laboratory)
Baiwu Zhang (Qualcomm)
Jingchi Zhang (Duke)
Xiaopeng Zhang (Qualcomm)
Jay Zhuo (Qualcomm)

Corresponding Author: Yung-Hsiang Lu, yunglu@purdue.edu